\documentclass[conference,onecolumn]{IEEEtran}
\usepackage{algorithm}
\usepackage{algorithmic}
\usepackage{float}
\usepackage{amsmath,amssymb}
\usepackage{bm}
\usepackage{hyperref}

\usepackage{graphicx}  
\usepackage{caption}   
\usepackage{pdflscape}
\begin{document}

\title{Enhanced Low-Density Region Exploration in Classifier-Guided Diffusion Models Through Modified Reverse Diffusion Sampling}

\author{\IEEEauthorblockN{Jagriti Singh}
\IEEEauthorblockA{Department of IT \\ IIIT Allahabad \\ rsi2023501@iiita.ac.in}
\and
\IEEEauthorblockN{Shekhar Verma}
\IEEEauthorblockA{Department of IT \\ IIIT Allahabad \\ sverma@iiita.ac.in}
\and
\IEEEauthorblockN{Muneendra Ojha}
\IEEEauthorblockA{Department of IT \\ IIIT Allahabad \\ muneendra@iiita.ac.in}
}

\maketitle  

\begin{abstract}
Diffusion models have emerged as state-of-the-art generative models for high-fidelity image synthesis, particularly in their classifier-free guided and classifier-guided forms. However, standard classifier guidance concentrates probability mass around high-density class mean, leading to poor coverage of rare samples in the tails of the class-conditional distributions. Recent work on diffusion-based tail sampling mitigates this by training an additional low-density–seeking classifier with a synthetic-vs-real discriminator, at the cost of additional networks and training. Also covering less possible regions by applying guidance at noise level by using one more diffusion model, which is a bad or weak version of 1st diffusion model. In parallel, a number of samplers and distillation techniques accelerate or refine diffusion sampling, but do not explicitly address long-tail coverage. 
We propose a purely sampling-time, density-aware extension of classifier-guided conditional diffusion model that targets low-density regions without any additional training. We have applied guidance at noisy images not on predicted noise like most diffusion models. Starting from a pretrained conditional diffusion model and classifier on ImageNet, we modify the guided reverse dynamics by steering trajectories toward low-confidence regions via the modified classifier gradient, and at each time step, we also guide the sampling process toward the predicted real image. 1st guidance helps explore low-probability samples, and 2nd guidance helps to generate samples to be close to the real data manifold. 
 The proposed sampler consistently improves ADM model recall at 64x64 resolution while maintaining a comparable FID, and with a 256x256 ADM model, we showed the results visually with different combinations of both guidance. We also showed that standard ADM classifier guidance, combined with predicted real image guidance, helps generate high perceptual quality samples with a 256x256 ADM model on ImageNet.
\end{abstract}


\section{Introduction}

Diffusion models have recently emerged as a powerful alternative to GANs for image synthesis, providing a robust and scalable generative framework based on iterative denoising of Gaussian noise~\cite{ho2020ddpm,song2019score,song2021sde}. In these models, a neural network is trained to predict either noise or the score (gradient of the log-density) at each time step, and sampling is performed by reversing a fixed or learned noising process. This training objective is numerically stable, scales well to large datasets, and yields state-of-the-art performance on high-resolution benchmarks. When combined with classifier guidance, diffusion models can be conditioned on semantic labels and achieve high perceptual quality on ImageNet~\cite{ddpm,imagenet}.

Despite this progress, generating rare or low-density examples within a class remains challenging. In classifier-guided diffusion, the sampling trajectory is steered by the gradient of the log-posterior of the classifier with respect to the current noisy sample. Increasing the guidance strength biases samples toward regions of high classifier confidence, effectively concentrating probability mass around high-density modes of the conditional data distribution~\cite{ddpm}. Naively driving the sampler toward the ``tail'' of the classifier distribution can push samples away from the empirical data manifold, resulting in low-fidelity images. To mitigate this, prior work has introduced additional mechanisms that explicitly constrain tail samples to remain realistic. For example, LDHF-DDPM\cite{rare} augments the diffusion model with a synthetic-vs-real discriminator loss, penalizing samples that deviate too far from the support of real images while still encouraging exploration of low-density regions in feature space.

In this work, the goal is to generate diverse samples that include both high-density and low-density samples, while preserving the high fidelity of classifier-guided diffusion model. Instead of training additional discriminators or tail-specific models, we keep both the pretrained conditional diffusion model ${p}_{\theta }(x_t|y)$ and the pretrained classifier ${p}_{\phi }(y|x_t)$ fixed, and modify only the sampling process. Concretely, we operate in the feature space induced by classifier-guided diffusion~\cite{ddpm} and alter the guidance at each diffusion step by (i) manipulating the classifier gradient towards a low density region and (ii) pushing the conditional mean ${p}_{\theta }(x_t|y)$ towards the predicted real sample $\hat{x}_0$ at each time step. By flipping the classifier gradient direction, we deliberately steer the sampler toward regions less favored by the classifier, thereby probing low-probability regions of the conditional distribution. Motivated by~\cite{rare}, we normalize the classifier gradient for controlled exploration of low-density regions with 64x64 resolution model, thereby reducing the likelihood of collapse into unrealistic images. For the 256x256 ADM model, we randomly selected 3 classes and generated 2 samples per class. Then perform a large number of experiments on them to understand the effect of guidance with different scales. This helps visualize the proposed guidance.

The resulting procedure can be viewed as a density-aware extension of classifier-guided diffusion model. Modified guidance is used to reliably sample from low-density regions, while the target-aware guidance steps intermittently help to be near target-density regions, increasing the diversity and fidelity of generated samples. We evaluate the method on ImageNet~\cite{imagenet}, using Fréchet Inception Distance (FID)~\cite{heusel2017fid} to assess fidelity and recall-style metrics derived from local manifold comparisons~\cite{kynkaanniemi2019pr,naeem2020prdc} to quantify how well the modified sampler recovers rare and low-density regions of the data distribution. Our experiments showed that the proposed sampling strategy can approximate the rare-sample behavior reported in~\cite{rare}, while maintaining competitive FID and using only a pretrained diffusion model and classifier.

\section{Related Work}
Diffusion models have recently become a dominant paradigm for high-fidelity image synthesis, with classifier-guided variants showing state-of-the-art performance on large-scale benchmarks such as ImageNet~\cite{ho2020ddpm,ddpm,imagenet}. This work focuses on methods that explicitly target low-density or long-tailed regions of the data distribution and on techniques that improve mode coverage and rare-sample generation.

Closest to this work, \cite{rare} modifies the sampling trajectory of a pretrained diffusion model to bias generation toward low-density regions identified in feature space, while preserving sample quality on ImageNet. Subsequent research on long-tailed diffusion further studies systematic biases of score-based models under imbalanced training distributions, proposing score-oriented calibration and class-aware mixing strategies to improve tail-class generation~\cite{zhang2024longtail}. Other approaches introduce explicit bias-correction solvers for diffusion-based image synthesis on long-tailed datasets, adjusting noise schedules or class-conditioning to enhance diversity in tail classes without sacrificing fidelity on head classes~\cite{ltbsolver2025}. Recent work on principled long-tailed generative modeling via diffusion formulates tail coverage as an optimization objective and combines reweighting with modified sampling to better match long-tail statistics~\cite{das2025longdiff}. In contrast to these methods, which alter training objectives or score calibration, our approach keeps both the diffusion model and classifier fixed and instead restructures the classifier-guided sampling process to simultaneously target high and low-density regions while maintaining the fidelity of the generated data.

A complementary line of work focuses on generating rare samples using GANs and other implicit models. Methods such as DeLiGAN introduce mixture-model parameterizations of the latent space to improve intra-class diversity under limited or diverse-small data regimes~\cite{deligan}. Boosting-based approaches, such as AdaGAN, refine mode coverage by iteratively reweighting training samples and fitting a sequence of generators, directly targeting missing modes and underrepresented regions~\cite{adagan}. More recently, rare-sample–specific frameworks explicitly aim at generating samples from the tail of the distribution: IL-GAN uses incremental learning to focus successive GAN updates on rare events while retaining high-level structure of common modes~\cite{ilgan}; \cite{lee2025divrare} formulates diverse rare sample generation with pretrained GANs as a multi-objective optimization problem in latent space, using normalizing flows to control rarity and diversity around a given reference; and \cite{khorram2024utlo} improves long-tailed class-conditional GANs by sharing low-resolution unconditional features between head and tail classes. Class-balancing regularizers leverage a pretrained classifier to encourage a GAN to generate a more balanced class distribution, improving performance on highly long-tailed datasets~\cite{rangwani2021cbgan}. Orthogonal to training-time modifications, sampling-based methods wrap a trained generator with corrective procedures: Metropolis–Hastings GANs construct an MCMC sampler around the GAN discriminator to better approximate the data distribution~\cite{turner2019mhgan}, and leverage-score–based sampling reweighs training data via ridge leverage scores to promote complete mode coverage, particularly in low-density regions~\cite{schreurs2021leverage}. These works highlight that rare-sample generation can be improved via either training-time reweighting or sampling-time corrections; our approach follows the latter philosophy, in the context of classifier-guided diffusion rather than GANs.

Evaluation methodology for rare-sample and mode-coverage studies commonly combines fidelity and coverage metrics. FID, based on Fréchet distance between Gaussian fits in Inception feature space, remains the standard measure of overall realism and distributional alignment~\cite{heusel2017fid}. Precision–recall–style metrics for generative models explicitly separate fidelity (precision) from coverage (recall) by comparing the local manifolds of real and generated data in a chosen feature space~\cite{kynkaanniemi2019pr}. Later work refines this into density and coverage (PRDC) metrics, improving robustness and interpretability of coverage estimates~\cite{naeem2020prdc}. The goal of diverse tail coverage.

Overall, prior work demonstrates that (i) diffusion models and GANs can achieve high fidelity but tend to under-represent low-density regions, (ii) training-time reweighting, architectural changes, or auxiliary samplers can mitigate this under-coverage, and (iii) recall-style metrics and perceptual feature spaces are essential to quantify improvements on rare samples. The proposed method differs by operating entirely at sampling time on a pretrained classifier-guided diffusion model, designing a family of guidance modifications.

\section{Methodology}
In the classifier-guided conditional diffusion model, the classifier guides the sampling process towards the desired class. However, the aim of this classifier guidance is to move towards the high-density regions within each class. So the final output from classifier-guided conditional diffusion models is the data from the high-density region. Our goal is to generate rare samples from the data distribution. For this, we modify the classifier guidance to move towards the tail of the distribution. In this way, we have generated rare samples, and also by adding guidance towards predicted real image $\hat{x}_0$ at each time step, we forced the generated data to be near the real data manifold.\\
We explore the diffusion model feature space by moving in the direction of the predicted real image $\hat{x}_0$ with scale $b$ and the opposite direction of the classifier mean of the desired class with scale $a$. We first perform experiments on a single class with different values of  $a$ and $b$. Increasing the value of $a$ yields samples with high recall but worsens fidelity, whereas increasing the value of $b$ yields samples with high fidelity but decreases recall. 

In the next subsections, we first recall the conditional DDPM formulation and classifier guidance, and then describe our density-aware modifications that target rare (low-density) intra-class samples while preserving high-fidelity generation. We also proposed another method to generate high perceptual quality images by combining standard ADM classifier ${p}_{\phi }(y|x_t)$ guidance with predicted real image classifier ${p}_{\phi }(y|\hat{x}_0)$ guidance.
\subsection{Diffusion-Based Model Preliminaries}
\subsubsection{Conditional Diffusion }

Let $x_0 \sim p_{\text{data}}(x)$ denote a clean image and $\{\beta_t\}_{t=1}^T$ a variance schedule with $\alpha_t = 1 - \beta_t$ and $\bar{\alpha}_t = \prod_{s=1}^t \alpha_s$. The forward diffusion process in DDPM~\cite{ho2020ddpm} is a fixed Markov chain
\begin{equation}
q(x_t \mid x_{t-1}) = \mathcal{N}\!\left(\sqrt{\alpha_t}\,x_{t-1}, \;\beta_t \mathbf{I}\right),
\end{equation}
which admits the closed-form marginal
\begin{equation}
q(x_t \mid x_0) = \mathcal{N}\!\left(\sqrt{\bar{\alpha}_t}\,x_0, \;(1 - \bar{\alpha}_t)\mathbf{I}\right), \quad t = 1,\dots,T.
\end{equation}

A class conditional diffusion model parameterizes a reverse-time Markov chain
\begin{equation}
p_\theta(x_{0:T}) = p(x_T)\prod_{t=1}^T p_\theta(x_{t-1} \mid x_t,y),
\end{equation}
where $p(x_T) = \mathcal{N}(0,\mathbf{I})$ and
\begin{equation}
p_\theta(x_{t-1} \mid x_t,y) = \mathcal{N}\!\left(\mu_\theta(x_t, t,y), \;\sigma_t^2 \mathbf{I}\right),
\end{equation}

where $y$ is a class label. So that class information is encoded directly in the network through label conditioning~\cite{ho2020ddpm,ddpm}. Conditional diffusion network outputs conditional noise $\varepsilon_\theta(x_t, t,y)$ and log variance ($\log \sigma_t^2$) at each time step.$\epsilon_{\theta}(x_t,t,y) \approx - \sigma_t \nabla_{x_t} \log p(x_t\mid y)$
Thus, even without any auxiliary classifier, the model provides an estimate of the conditional distribution $p(x_t \mid y)$.
$\sigma_t^2$ is either tied to the forward noise schedule $\{\beta_t\}$ or parameterized as an additional output of the network~\cite{ho2020ddpm,nichol2021improved}. In our experiments, we follow the learned-variance formulation of~\cite{nichol2021improved} and use the predicted log variances during sampling. In the commonly used noise-prediction parameterization, the mean can be written as
\begin{equation}
\mu_\theta(x_t, t,y) = \frac{1}{\sqrt{\alpha_t}}
\left(
x_t - \frac{\beta_t}{\sqrt{1 - \bar{\alpha}_t}}\,\varepsilon_\theta(x_t, t,y)
\right),
\end{equation}
where $\varepsilon_\theta(x_t, t,y)$ approximates the conditional gaussian noise added at time $t$.

With the help of the reparameterization trick  $ \hat x_0$ (predicted real image) can be calculated directly at any time step t as below
\begin{equation}
\hat{x}_0 = \frac{x_t - \sqrt{1 - \bar{\alpha}_t}\,\epsilon_\theta(x_t, t,y)}{\sqrt{\bar{\alpha}_t}}
\label{eq:x_0}
\end{equation}

\subsubsection{Guided Conditional Diffusion }

In the classifier-guided conditional diffusion model, two trained neural networks are used. One is a conditional diffusion model $ {p}_{\theta }(x_t|y)$, and the other is a classifier model ${p}_{\phi }(y|x_t)$ trained on noisy images. The reverse process for sampling uses the classifier score along with the conditional mean as follows.
\begin{equation}
p_\theta(x_{0:T} \mid y) = p(x_T)\prod_{t=1}^T p_\theta(x_{t-1} \mid x_t, y, \nabla_{x_t}\log p_\phi(y \mid x_t)),
\end{equation}
Class information is encoded directly in the network via label conditioning, enhanced by the gradient information from the separate classifier. Calculating  $p_\theta(x_{t-1} \mid x_t, y)$ depends on both networks.

\begin{equation}
p_\theta(x_{t-1} \mid x_t, y)
= \mathcal{N}\!\left({\mu}_g(x_t, t, y),\; \sigma_t^2 \mathbf{I}\right),
\label{e}
\end{equation}
$\mu_g$ is a guided mean, obtained by combining the conditional mean and the classifier gradient.
Our objective is to generate diverse samples within each class, explicitly covering both high-density and low-density regions of the class-conditional distribution. Standard classifier guidance promotes modes where $p_\phi(y \mid x_t)$ is high, thereby undersampling rare intra-class samples. To target such rare regions, we modify the guidance term in a density-aware manner.
   
\subsection{Proposed Density-aware Guided Conditional Diffusion Model }
In classifier guided ADM model target distribution $\tilde {p}_{\theta }(x_t|y)$ is proportional to conditional diffusion model ${p}_{\theta }(x_t|y)$ and classifier model ${{p}_{\phi }(y|x_t)}$. Both models are different neural networks. Using the first conditional diffusion network, the mean is calculated at each time step and is guided by the classifier model with guidance scaler value $s$.
\begin{equation}
\tilde {p}_{\theta }(x_t|y) \propto {p}_{\theta }(x_t|y) {{p}_{\phi }(y|x_t)}^s
\end{equation}

In our setup, we add one more guidance to the conditional model. New guidance is based on the predicted real image $\hat{x}_0$ at each time step. Below is the new proposed target distribution $\tilde{p}_{\theta}(x_t|y)$ for high perceptual-quality image generation, and to generate low-probability samples by changing the direction of classifier $p_{\phi}(y|x_t)$.
\begin{equation}
\tilde {p}_{\theta }(x_t|y) \propto {p}_{\theta }(x_t|y) {{p}_{\phi }(y|x_t)}^a{{p}_{\phi }(y|\hat{x}_0)}^b
\end{equation}

$a$ and $b$ are positive scalar values for generating high perceptual quality images. We used a pretrained classifier trained on noisy images $x_t$ to guide the sampling process towards the predicted real image $\hat{x}_0$. The classifier is trained on $x_t$, and the predicted real image $\hat{x}_0$ is a less noisy version of $x_t$ as mentioned in the equation \eqref{eq:x_0}. That's why the classifier worked on $\hat{x}_0$, and increasing this guidance by increasing the value of $b$ did not smooth the generated image, unlike classifier guidance on $x_t$ with a higher guidance scaler value, as mentioned in ~\cite{ddpm}. Guidance towards the target class helps the generated sample achieve finer details of the target class, as shown in the experiment and results sections. We also used these 2 guidance to incorporate features of 2 different classes. For this, we have decided 4 cases to show the effect of guidance more clearly. 

\subsubsection{Similar features opposite direction}
In this case, we fixed the conditional diffusion model base class $y_{base}$ and changed the target class for the classifier to the class with similar features to the base class, i.e., ${p}_{\theta }(x_t|y_{base})$. We have used both guidance based on the classifier, one in the direction of the base class $y_{base}$ and the other in the direction of the target class $y_{target}$. Classifier guidance towards target class $\nabla_{x_t} \log{p}_{\phi }(y_{target}|x_t)$ helps conditional diffusion model to have features of target class and Classifier guidance towards base class $\nabla_{\hat{x}_0} \log{p}_{\phi }(y_{base}|\hat{x}_0)$ helps conditional diffusion model to move towards the base class.

\begin{equation}
\tilde {p}_{modified }(x_t|y_{base->target}) \propto {p}_{\theta }(x_t|y_{base}) {{p}_{\phi }(y_{target}|x_t)}^a{{p}_{\phi }(y_{base}|\hat{x}_0)}^b 
\end{equation}

\begin{figure}[H]
    \centering
    \includegraphics[width=\textwidth]{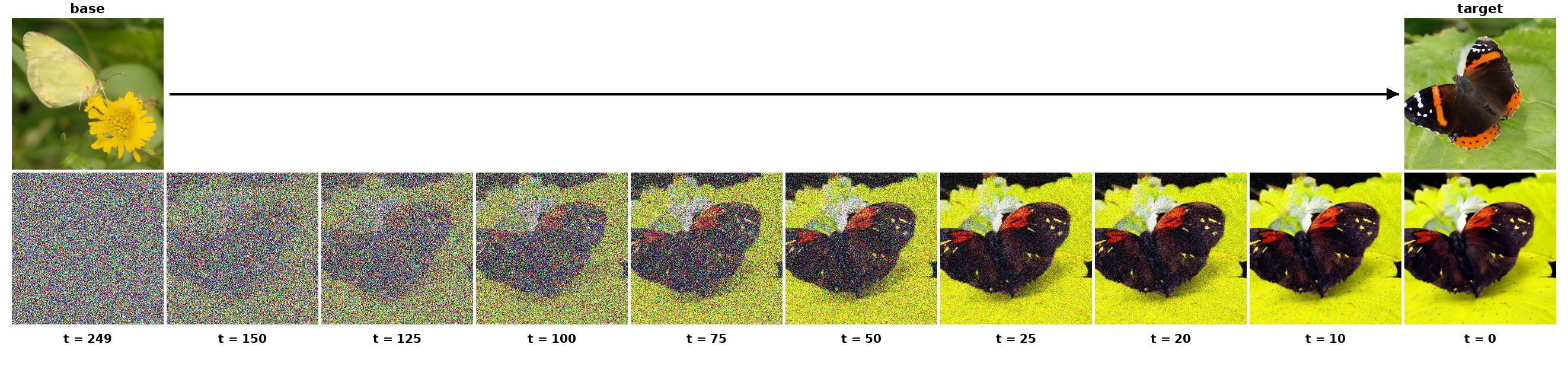}
    \caption{ Generation of image from base class with class ID 325, towards target class with class ID 321. With one classifier model ${p}_{\phi }(y_{target}|x_t)$ guidance towards the target class and other ${p}_{\phi }(y_{base}|\hat{x}_0)$  based guidance towards the base class. Value of $a $ is $= 63$ and $b $ is $= 1$.}
    \label{fig:so}
\end{figure}

As shown in the figure ~\ref{fig:so}. Both base and target classes have similar features, which helps the generation process produce diverse samples that inherit properties of both classes and achieve high perceptual quality, compared to dissimilar base and target classes. Scalar values $a$ and $b$ guide the generation process to explore different regions of the mixture of feature space. Both base and target images have been generated by the standard ADM-G 256x256 Model~\cite{ddpm} with the same seed, but with different class labels (325, 321). 

\subsubsection{Similar features same direction}
In this case, like the previous one, we fixed the conditional diffusion model base class $y_{base}$  and changed the target class for the classifier, with the class that has similar features to the base class ${p}_{\theta }(x_t|y_{base})$. Again, we have used both guidance based on the classifier, but both in the direction of the target class $y_{target}$. Both classifier-based guidance towards target class $\nabla_{x_t} \log{p}_{\phi }(y_{target}|x_t)$ and $\nabla_{\hat{x}_0} \log{p}_{\phi }(y_{target}|\hat{x}_0)$ help the conditional diffusion model to have features of the target class. 
\begin{equation}
\tilde {p}_{modified }(x_t|y_{base->target}) \propto {p}_{\theta }(x_t|y_{base}) {{p}_{\phi }(y_{target}|x_t)}^a{{p}_{\phi }(y_{target}|\hat{x}_0)}^b 
\end{equation}

\begin{figure}[H]
    \centering
    \includegraphics[width=\textwidth]{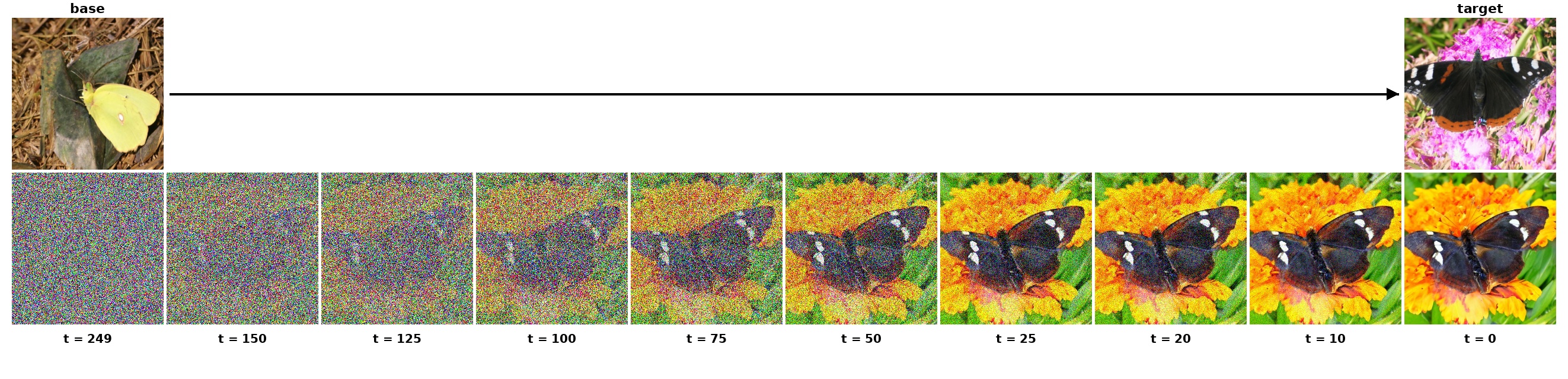}
    \caption{Generation of image from base class with class ID 325, towards target class with class ID 321. With both classifier-based guidance towards the target class.Value of $a $ is $= 99$ and $b $ is $= 99$.}
    \label{fig:ss}
\end{figure}

As shown in the figure ~\ref{fig:ss}. Both base and target classes have similar features, which helps the generation process produce diverse samples that inherit properties of both classes and achieve high perceptual quality, compared to dissimilar base and target classes. Scalar values $a$ and $b$ guide the generation process to explore different regions of the mixture of feature space. Both base and target images have been generated by the standard ADM-G 256x256 Model~\cite{ddpm} with the same seed, but with different class labels (325, 321). 

\subsubsection{Dissimilar features opposite direction}
In this case, we fixed the conditional diffusion model base class $y_{base}$ and changed the target class for the classifier to the class with non-similar features to the base class, i.e., ${p}_{\theta }(x_t|y_{base})$. We have used both guidance strategies based on the classifier: one toward the base class $y_{base}$ and the other toward the target class $y_{target}$. Classifier guidance towards target class distribution $\nabla_{x_t} \log{p}_{\phi }(y_{target}|x_t)$ helps conditional diffusion model to have features of target class and Classifier guidance towards base class $\nabla_{\hat{x}_0} \log{p}_{\phi }(y_{base}|\hat{x}_0)$ helps conditional diffusion model to move towards the base class.

\begin{figure}[H]
    \centering
    \includegraphics[width=\textwidth]{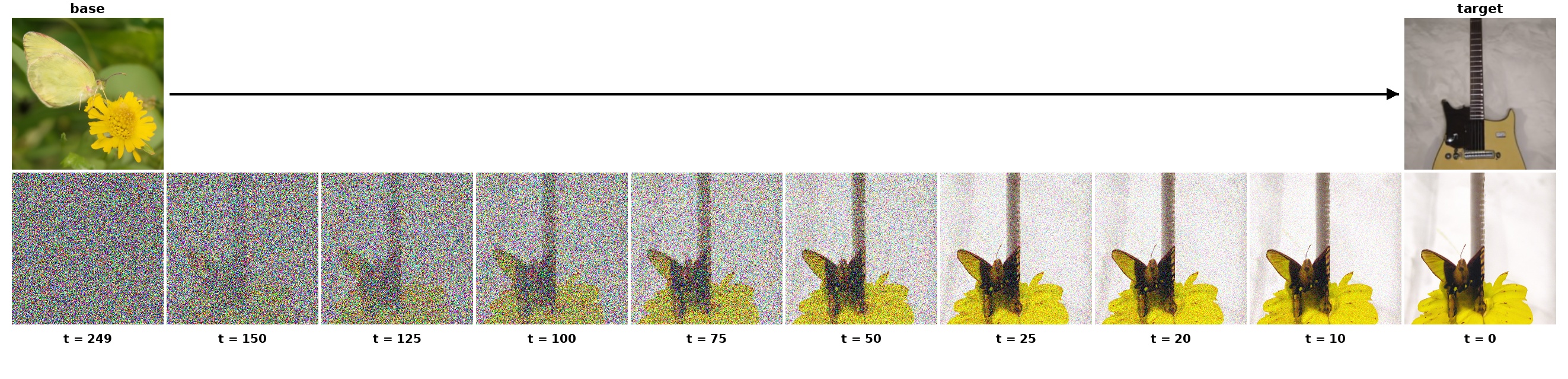}
    \caption{Generation of image from base class with class ID 325, towards target class with class ID 546. With one classifier model ${p}_{\phi }(y_{target}|x_t)$ guidance towards the target class and other ${p}_{\phi }(y_{base}|\hat{x}_0)$ based guidance towards the base class.Value of $a $ is $= 10$ and $b $ is $= 13$.}
    \label{fig:oo}
\end{figure}

As shown in the figure ~\ref{fig:oo}. Both base and target classes have non-similar features, which helps the generation process produce diverse samples that have properties of both classes. Scalar values $a$ and $b$ guide the generation process to explore different regions of the mixture of feature space. Both base and target images have been generated by the standard ADM-G 256x256 Model~\cite{ddpm} with the same seed, but with different class labels (325,546).

\subsubsection{Dissimilar features Same direction}
In this case, like the previous one, we fixed the conditional diffusion model base class $y_{base}$  and changed the target class for the classifier, with the class that has non-similar features to the base class ${p}_{\theta }(x_t|y_{base})$. Again, we have used both guidance based on the classifier,  but both in the direction of the target class $y_{target}$. Both classifier-based guidance towards target class $\nabla_{x_t} \log{p}_{\phi }(y_{target}|x_t)$ and $\nabla_{\hat{x}_0} \log{p}_{\phi }(y_{base}|\hat{x}_0)$ help the conditional diffusion model to have features of the target class.

\begin{figure}[H]
    \centering
    \includegraphics[width=\textwidth]{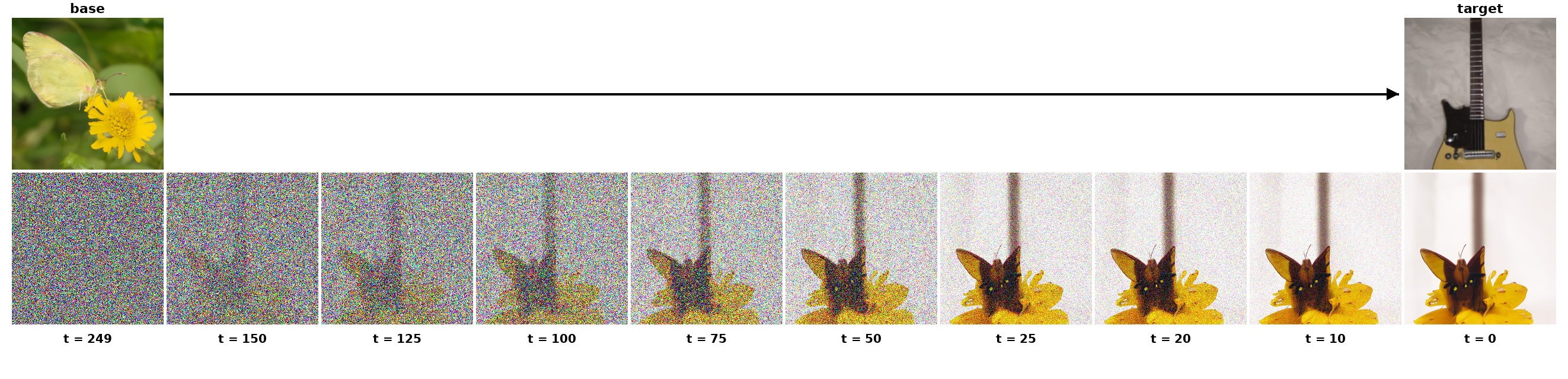}
    \caption{Generation of image from base class with class ID 325, towards target class with class ID 321. With both classifier-based guidance towards the target class.Value of $a $ is $= 10$ and $b $ is $= 1$.}
    \label{fig:same546}
\end{figure}

As shown in the figure ~\ref{fig:same546}. Both base and target classes have non-similar features, which helps the generation process produce diverse samples that have properties of both classes. Scalar values $a$ and $b$ guide the generation process to explore different regions of the mixture of feature space. Both base and target images have been generated with the same seed by the standard ADM-G 256x256 Model~\cite{ddpm}, and only with different class labels (325,546).\\

\begin{figure}[H]
    \centering
    \includegraphics[width=\textwidth]{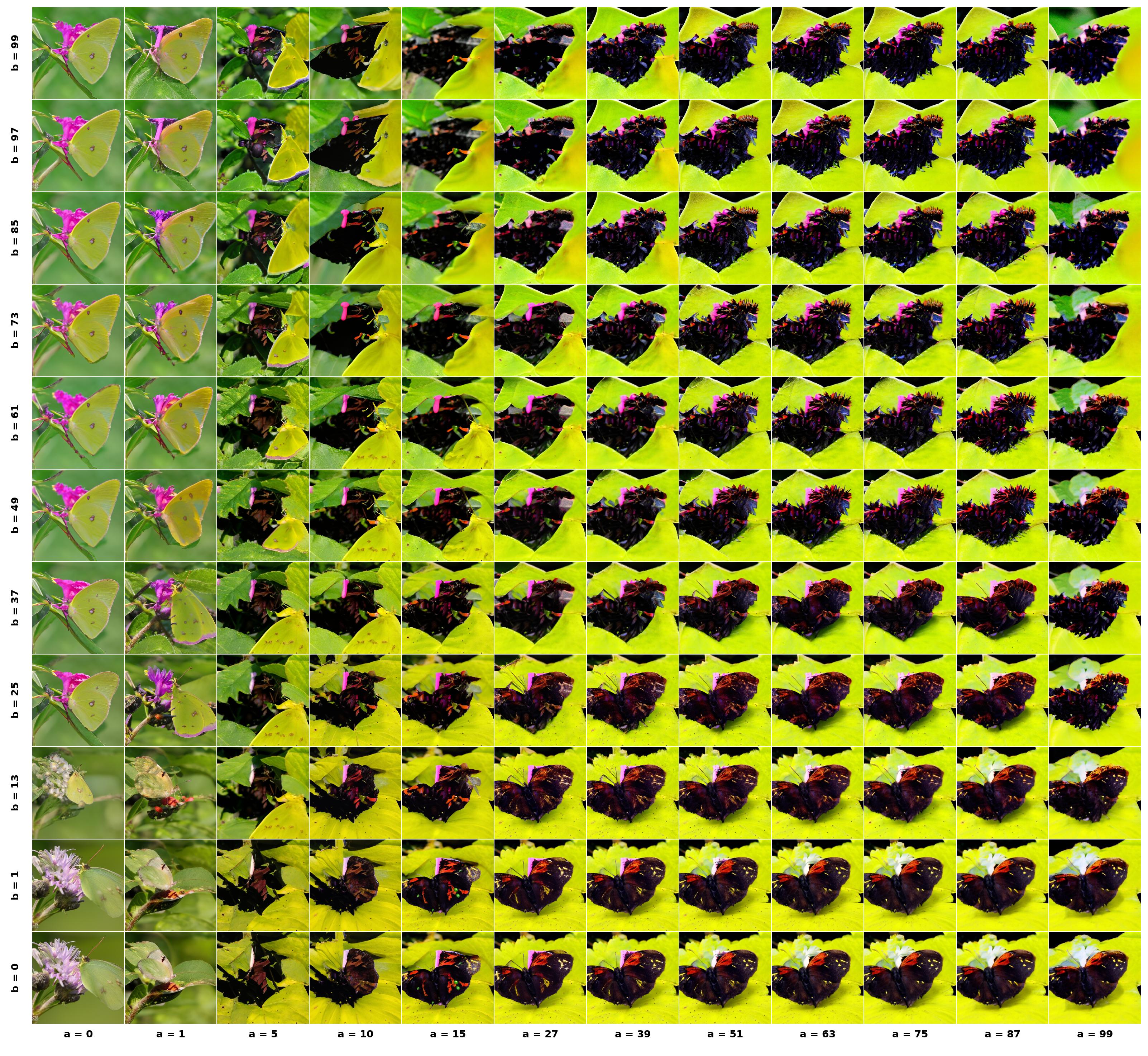}
    \caption{Generation of images from base class with class ID 325, towards target class with class ID 321. With one classifier model ${p}_{\phi }(y_{target}|x_t)$ guidance towards the target class and other ${p}_{\phi }(y_{base}|\hat{x}_0)$  based guidance towards the base class. With different values of $a $ and $b $.}
    \label{fig:s1}
\end{figure}

\begin{figure}[H]
    \centering
    \includegraphics[width=\textwidth]{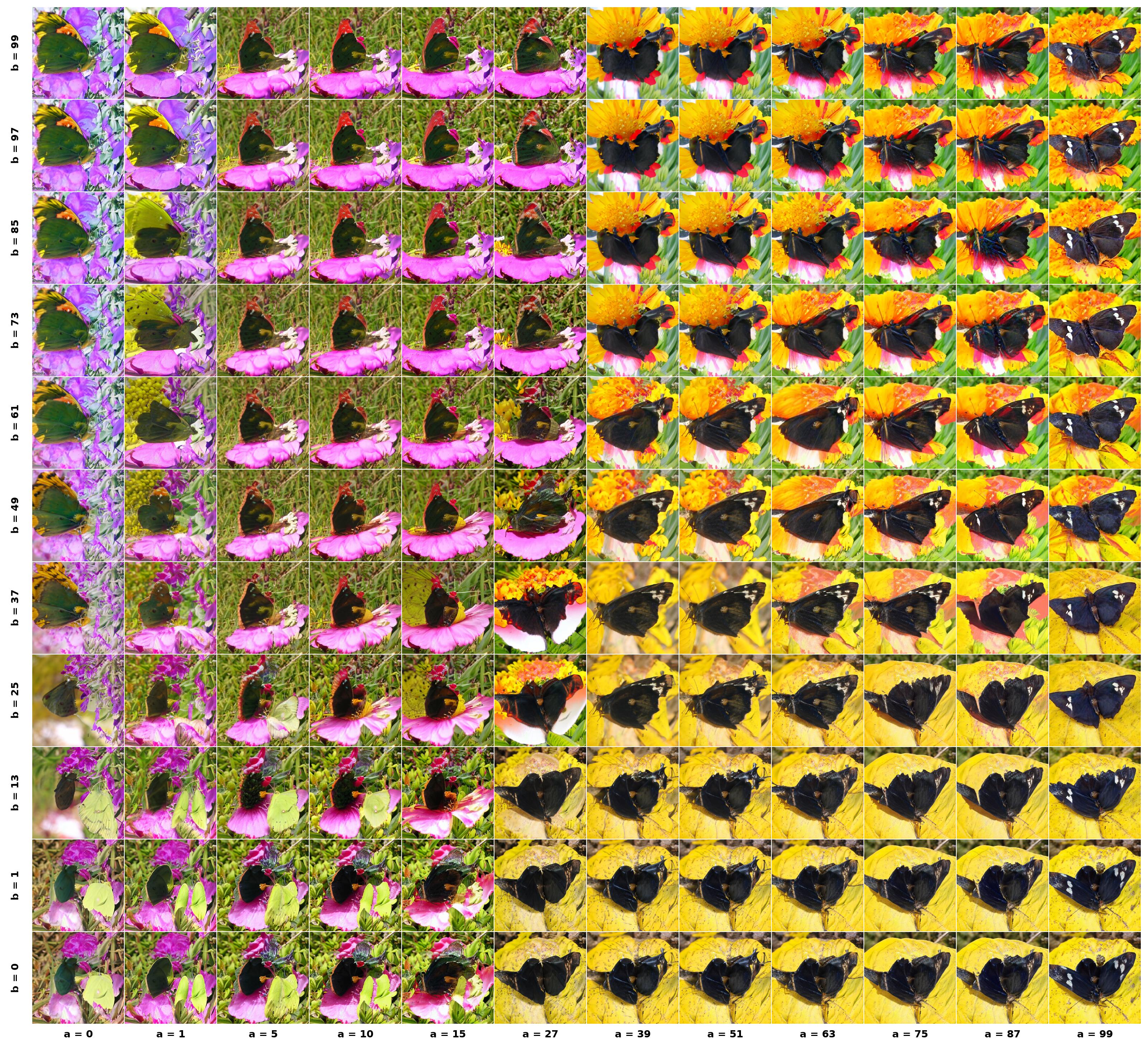}
    \caption{Generation of images from base class with class ID 325, towards target class with class ID 321. With both classifier-based guidance towards the target class. With different values of $a $ and $b $.}
    \label{fig:s3}
\end{figure}

\begin{figure}[H]
    \centering
    \includegraphics[width=\textwidth]{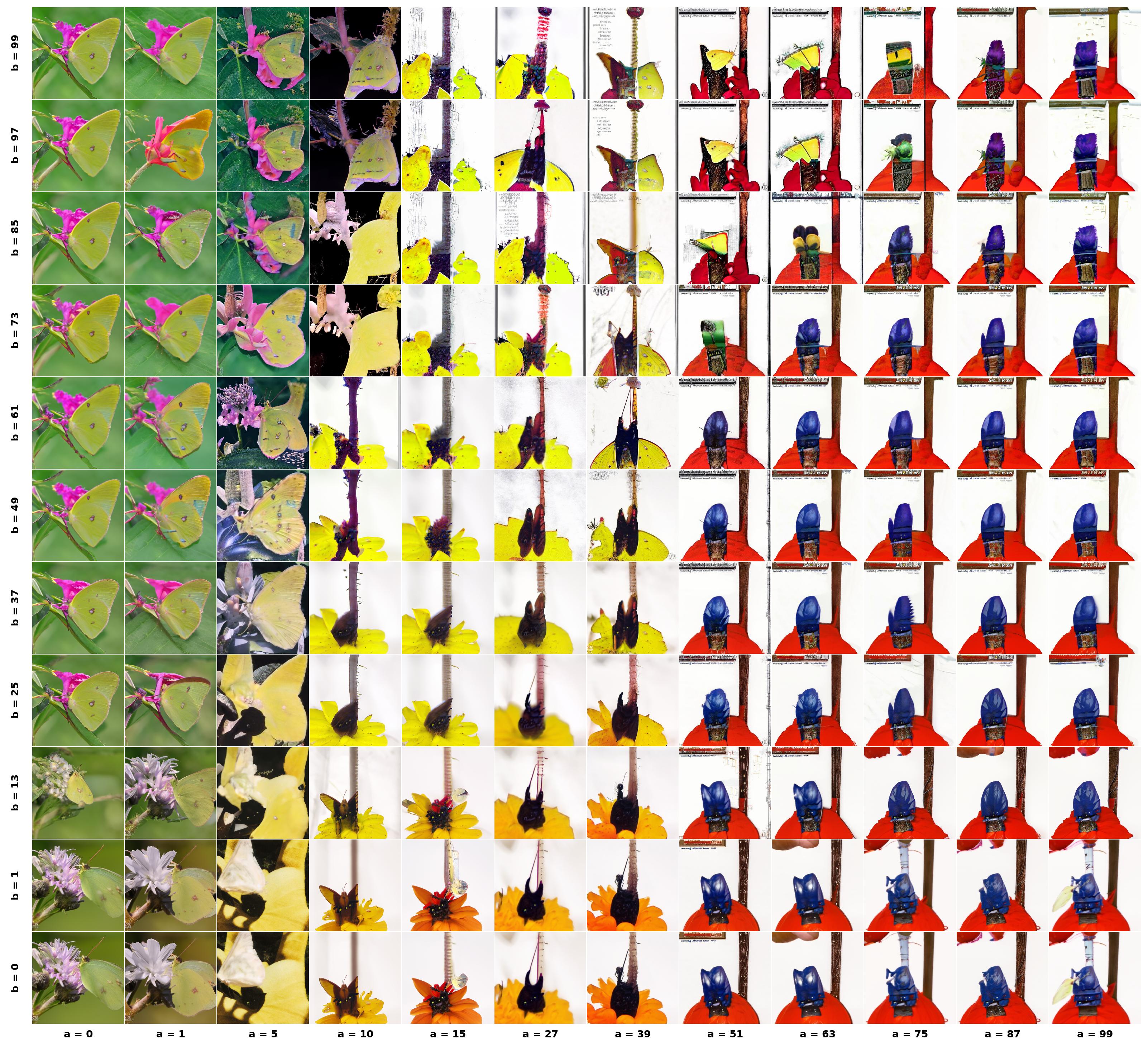}
    \caption{Generation of images from base class with class ID 325, towards target class with class ID 546. With one classifier model ${p}_{\phi }(y_{target}|x_t)$ guidance towards the target class and other ${p}_{\phi }(y_{base}|\hat{x}_0)$ based guidance towards the base class. With different values of $a $ and $b $.}
    \label{fig:s2}
\end{figure}

\begin{figure}[H]
    \centering
    \includegraphics[width=\textwidth]{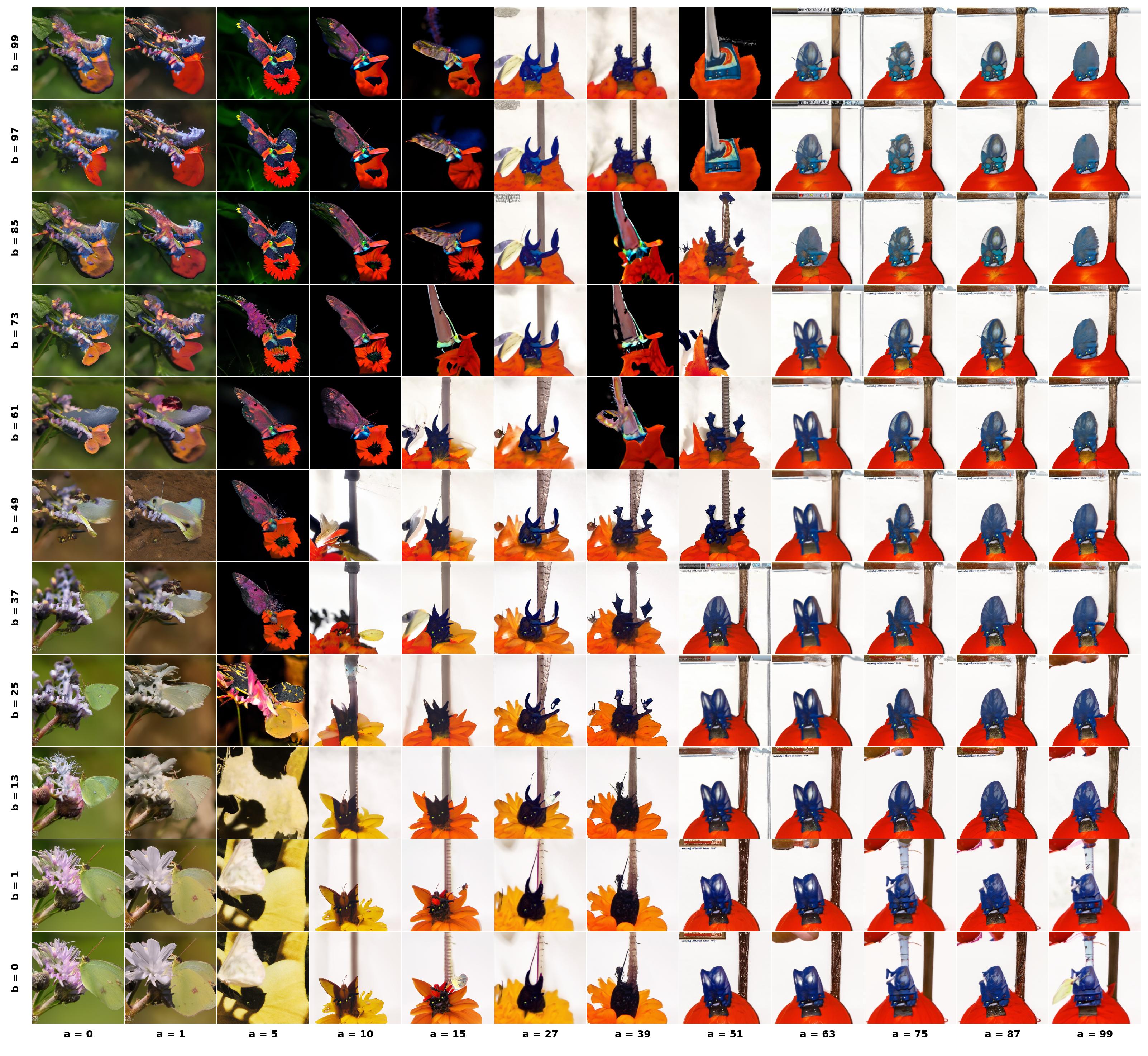}
    \caption{Generation of images from base class with class ID 325, towards target class with class ID 321. With both classifier-based guidance towards the target class. With different values of $a $ and $b $.}
    \label{fig:s4}
\end{figure}

The above 4 cases clearly showed that guidance can help to generate diverse samples. For above mentioned all 4 cases, the modified distribution images for different values of $a$ and $b$ have been shown in the figure ~\ref{fig:s1},~\ref{fig:s3}, ~\ref{fig:s2}, and ~\ref{fig:s4} respectively. During the generation of base-class images with one target class, it is perceptually evident that target-class guidance, distinct from the base class, helps move the conditional diffusion model's mean away from the base-class distribution at each step. But guidance in the opposite direction of the base class mean ${p}_{\phi }(y_{base}|x_t)$ helps to capture features of all the rest classes(other than the base class), but is not perceptually visible. In our proposed methods for generating diverse samples, we have modified the classifier guidance to move in the opposite direction of the base class, thereby moving the conditional diffusion mean away from the base class mean and guidance $\nabla_{\hat{x}_0} \log{p}_{\phi }(y_{base}|\hat{x}_0) $ helped generation process to be near real data distribution, leads to explore rare region of base class distribution. During low possible region exploration, proposed distribution  $\tilde {p}_{\theta }(x_t|y_{base})$ is
\begin{equation}
\tilde {p}_{\theta }(x_t|y_{base}) \propto {p}_{\theta }(x_t|y_{base}) {{p}_{\phi }(y_{base}|x_t)}^{-a}{{p}_{\phi }(y_{base}|\hat{x}_0)}^b
\end{equation}
${{p}_{\phi }(y_{base}|x_t)}^{-a}$ guides the generated sample towards the mean of the non-desirable classes, and $\nabla_{\hat{x}_0} \log{p}_{\phi }(y_{base}|\hat{x}_0) $ guidance moves the generated sample towards the base class distribution. A combination of both guidance generates diverse samples from low density region with high fidelity. 

During sample generation by the classifier-guided diffusion model, the classifier score for the noisy image is computed at each time step. For the class label $y_{base}$, the score is proposed as $p_\phi(y_{base} \mid x_t)$. 

\begin{equation}
p_\phi(y_{base} \mid x_t) + p_\phi(\hat y_{base} \mid x_t) = 1
\end{equation}
\begin{equation}
-p_\phi(y_{base} \mid x_t)  = p_\phi(\hat y_{base} \mid x_t) - 1
\end{equation}
\begin{equation}
\tilde{g}_t = -\nabla_{x_t}  p_\phi(y_{base} \mid x_t)  = \nabla_{x_t}  p_\phi(\hat y_{base} \mid x_t) 
\end{equation}
Where $\hat y_{base}$ is the rest of the classes other than $y_{base}$. Moving in the opposite direction of the mean of the classifier is equivalent to moving in the direction of the mean of the other classes. This $\tilde{g}_t$ pushed the sampling process away from the desired class, but due to variance, it cannot jump high into other classes. This guidance $\tilde{g}_t$ pushes the conditional mean of the diffusion model towards the low possible region of class $y_{base}$, because guidance is limited by variance, and the conditional mean pushes the generating samples towards the class $y_{base}$ data distribution along with $\nabla_{\hat{x}_0} \log{p}_{\phi }(y_{base}|\hat{x}_0) $ guidance.
 We can also apply normalization, which prevents the gradient magnitude from dominating the diffusion dynamics at any particular step. We then constructed a rare-sample–oriented mean

\begin{equation}
p_\theta(x_{t-1} \mid x_t, y_{base})
= \mathcal{N}\!\left({\tilde \mu}_\theta(x_t, t, y_{base}),\; \sigma_t^2 \mathbf{I}\right),
\label{eqq}
\end{equation}

which is used to modify the mean according to the variance-weighted conditioning strategy of~\cite{ddpm}, with modifications to allow samples from low-density regions. For generating 64x64 ImageNet samples, we used normalized gradient to improve the performance as mentioned in ~\cite{rare}. During 256x256 ImageNet sample generation, with a higher guidance scaler value and without normalization, better samples were generated. In our implementation, the classifier-guided low-density aware mean for the 64x64 ImageNet sample generation is
\begin{equation}
\tilde{\mu}_g(x_t, t, y_{base})
= \mu_\theta(x_t, t,y_{base}) \;+\; a\;*\Sigma_t ,*  \;\mathit{Normalize}(\tilde g_t),
\label{eq:guidance-variance-weighted}
\end{equation}
i.e., the conditional mean is shifted by the product of the model log variance($\Sigma_t$) and the modified classifier gradient and scaled by a positive scalar value $a$ to explore low-density regions. Then, we have added predicted real-image guidance $\nabla_{\hat{x}_0} \log{p}_{\phi }(y_{base}|\hat{x}_0) $  at time step t to help the generated sample remain close to the real data manifold. For this, we have computed the base distribution-aware guidance ($g_{\hat{x}_0}$) as below.

\begin{equation}
g_{\hat{x}_0} = \Sigma_t *  \mathit{Normalize}(\nabla_{\hat{x}_0}\log p_\phi(y_{base} \mid \hat{x}_0))
\end{equation}
$\hat{x}_0$ is calculated with the help of equation \eqref{eq:x_0}. $g_{\hat{x}_0}$ is scaled by $b$ to explore feature space of ADM model. For 256x256 ImageNet sample generation, with higher scaler values ($a,b$) and without normalization, we obtain better results. Multiplying guidance by variance $\Sigma_t$ helps explore the feature space within the feasible region; otherwise, using guidance directly yields unpredictable samples. 
We perform experiments over different values of $a$ and $b$ on a single class (ID 300) to understand the effect of both guidance, so we can generate diverse, high-fidelity images. We find a better tradeoff with $a = 0.5$ and $b = 0.5$ for the 64x64 ImageNet dataset. The corresponding sampling update can be written as

\begin{equation}
\tilde{\mu}_\theta(x_t, t, y_{base})
=  \tilde{\mu}_g(x_t, t, y_{base}) + b g_{\hat{x}_0},
\label{eq:guided-sample-step_new}
\end{equation}
\begin{equation}
x_{t-1}
= \tilde{\mu}_\theta(x_t, t, y_{base})
  + \tilde \Sigma_t\,\varepsilon_t,
\label{eq:guided-sample}
\end{equation}
\begin{equation}
\tilde \Sigma_t = \exp\!\bigl(\tfrac{1}{2}\log \Sigma_t\bigr),
\qquad
\varepsilon_t \sim \mathcal{N}(0,\mathbf{I}).
\label{eq:guided-noise}
\end{equation}

where $\log \Sigma_t $(log of variance) is obtained from the ADM model~\cite{ddpm}, and no additional noise is added at the final step $t=0$. Equations~\eqref{eq:guidance-variance-weighted}--\eqref{eq:guided-noise} correspond to the practical implementation of this proposed method. Through experimentation, we found that increasing $a$ generates diverse samples, and increasing $b$ generates high-fidelity samples, with a threshold beyond which further increases yield less diverse, lower-fidelity images. The threshold values for both $a$ and $b$ depend on each other. For high value of $b$, the threshold value for $a$ is high and vice versa. We found a better trade-off between diversity and fidelity at $a = 0.5$ and $b = 0.5$.  In this work, we modify this guidance term to deliberately explore low-density regions within each class. Sampling algorithm is mentioned in algorithm ~\ref{alg:diffusion_sampling}. In algorithm ~\ref{alg:diffusion_sampling}, $a$ and $b$ have positive scalar values.

\begin{algorithm}[!b]
\caption{ Density Aware Guided Conditional Diffusion Model.}
\label{alg:diffusion_sampling}
\begin{algorithmic}[1]
\REQUIRE Class Label $y_{base}$,  Conditional Diffusion Model $(\mu_\theta(x_t,t,y_{base}), \Sigma_\theta(x_t,t,y_{base}),\hat{x}_0)$, Classifier $p_\phi(y_{base}\mid x_t)$, Diversity Controller $a$, Fidelity Controller $b$
\ENSURE Generated Sample $x_0$ after $T$ Denoising Steps

\STATE Initialize $\bm{x}_T \sim \mathcal{N}(\bm{0}, \bm{I})$
\FOR{$t = T, T-1, \dots, 1$}
    \STATE $\mu, \Sigma \gets \mu_\theta(x_t,t,y_{base}), \Sigma_\theta(x_t,t,y_{base})$
    \STATE  $\mu_{g}  \gets \mu - a * \Sigma  \mathit{Normalize}( \nabla_{x_t} \log p_\phi(y_{base}\mid x_t))$
    \STATE $\tilde\mu_{}  \gets  \mu_g + b *  \Sigma  \mathit{Normalize}(\nabla_{\hat{x}_0}\log p_\phi(y_{base} \mid \hat{x}_0)) $
    \IF{$t > 1$}
        
        \STATE Update: $\bm{x}_{t-1} \gets \mathcal{N}\!\big(\tilde\mu_ , \Sigma\big)$ 
    \ELSE
        \STATE $\bm{x}_0 \gets \bm{\tilde\mu}$
    \ENDIF
\ENDFOR
\RETURN $\bm{x}_0$ 
\end{algorithmic}
\end{algorithm}

\begin{algorithm}[!b]
\caption{ High Perceptual Quality Guided Conditional Diffusion Model.}
\label{alg:diffusion_sampling_2}
\begin{algorithmic}[1]
\REQUIRE Class Label $y_{base}$,  Conditional Diffusion Model $(\mu_\theta(x_t,t,y_{base}), \Sigma_\theta(x_t,t,y_{base}),\hat{x}_0)$, Classifier $p_\phi(y_{base} \mid x_t)$, Fidelity Controller $a,b$
\ENSURE Generated Sample $x_0$ after $T$ Denoising Steps

\STATE Initialize $\bm{x}_T \sim \mathcal{N}(\bm{0}, \bm{I})$
\FOR{$t = T, T-1, \dots, 1$}
    \STATE $\mu, \Sigma \gets \mu_\theta(x_t,t,y_{base}), \Sigma_\theta(x_t,t,y_{base})$
    \STATE  $\mu_{g}  \gets \mu + a *  \Sigma   ( \nabla_{x_t} \log p_\phi(y_{base}\mid x_t))$
    \STATE $\tilde\mu_{}  \gets  \mu_g + b *  \Sigma   (\nabla_{\hat{x}_0}\log p_\phi(y_{base} \mid \hat{x}_0)) $
    \IF{$t > 1$}
        
        \STATE Update: $\bm{x}_{t-1} \gets \mathcal{N}\!\big(\tilde\mu_ , \Sigma\big)$ 
    \ELSE
        \STATE $\bm{x}_0 \gets \bm{\tilde\mu}$
    \ENDIF
\ENDFOR
\RETURN $\bm{x}_0$ 
\end{algorithmic}
\end{algorithm}

In all experiments, we used the publicly available classifier guided diffusion model from~\cite{ddpm}, generated images, and evaluated the resulting samples using FID~\cite{heusel2017fid} for fidelity and recall~\cite{kynkaanniemi2019pr} to quantify the recovery of rare, low-density regions in the class-conditional feature space.

\section{Experiments and Results}

\subsection{Evaluation on ImageNet 64$\times$64}

All experiments are conducted on the ImageNet dataset~\cite{imagenet} with $64\times 64$ resolution, with the goal of jointly evaluating image fidelity and coverage of low-density regions. We follow the evaluation protocol of the OpenAI guided-diffusion codebase~\cite{ddpm}: Fréchet Inception Distance (FID) is computed using Inception-v3 features~\cite{szegedy2016inception,heusel2017fid} against the standard ImageNet 64$\times$64 statistics. For calculating FID and recall, we have used the method described in~\cite {kynkaanniemi2019pr}, which quantifies how well the generated manifold covers the support of the real data distribution.

For ADM  and ADM-G ~\cite{ddpm}, LDHF-DDPM~\cite{rare}, and all variants of our proposed method, we generate samples with the same diffusion schedule and number of reverse steps as in the original ADM implementation, and evaluate FID and recall with the official OpenAI evaluator. For other generative baselines---progressive distillation (PD), consistency distillation (CD), DDIM, DPM-Solver, consistency training (CT), improved CT (iCT), StyleGAN2, and StyleGAN2+GANdance---we report the ImageNet 64$\times$64 FID and recall values from the GANdance study~\cite{kim2024gandance}, which uses the same metrics and feature extractor.

\subsubsection{Comparison on ImageNet 64$\times$64}

Table~\ref{tab:15} summarizes FID and recall on the ImageNet statistics, aggregated over all classes. One-step and few-step distilled diffusion models (PD, CD, CT, iCT)~\cite{salimans2022progressivedistillation,song2023consistency,song2024ict} provide fast sampling but generally lie off the best fidelity--recall frontier: for example, PD and CD achieve moderate recall but noticeably worse FID than ADM. High-order samplers such as DDIM~\cite{song2021ddim} and DPM-Solver~\cite{lu2022dpmsolver} improve either FID or recall relative to ADM-G, but none dominate ADM on both metrics simultaneously. StyleGAN2 and StyleGAN2+GANdance~\cite{karras2020stylegan2,kim2024gandance} exhibit competitive recall for GAN-based models but trail diffusion-based approaches in FID on this benchmark.

LDHF-DDPM~\cite{rare} substantially increases recall compared to ADM and ADM-G by explicitly biasing sampling toward low-density regions using an auxiliary classifier and a synthetic-vs-real discriminator. This gain in coverage, however, comes at the expense of FID, reflecting the difficulty of populating rare modes while remaining close to the real data manifold.

The proposed density-aware guided conditional diffusion model forms a smooth family of operating points parameterized by $(a,b)$. As $a$ increases, FID worsens, while recall increases. Several configurations offer good trade-offs between FID and recall. We have only mentioned 2 configuration results. Settings such as $(a,b)=(0.5,0.5)$  achieve FID substantially better than LDHF-DDPM with essentially slightly lower recall, demonstrating that our sampling-time modification can recover a large fraction of LDHF-style tail coverage without any additional networks. 
Overall, the results in Table~\ref{tab:15} suggest that the proposed sampler offers a flexible and competitive trade-off in terms of values; it bridges the gap between high-fidelity but mode-biased diffusion baselines and tail-focused methods like LDHF-DDPM, while requiring only a modified guidance rule on top of a fixed, pretrained conditional diffusion model and classifier.

\begin{table}[H]
\centering
\renewcommand{\arraystretch}{1.7}
\begin{tabular}{l c c}
\hline
Method & FID $\downarrow$ & Recall $\uparrow$ \\
\hline
PD (2-step)~\cite{salimans2022progressivedistillation}              &  8.95  & 0.65 \\
CD (2-step)~\cite{song2023consistency}                               &  4.70  & 0.64 \\
PD (1-step)~\cite{salimans2022progressivedistillation}              & 15.39  & 0.62 \\
CD (1-step)~\cite{song2023consistency}                               &  6.20  & 0.63 \\
ADM-G~\cite{ddpm}                                                    &  5.4752 & 0.5292 \\
ADM~\cite{ddpm}                                                      &  2.0779 & 0.6291 \\

DDIM~\cite{song2021ddim}                                             & 13.70  & 0.56 \\
DPM-Solver~\cite{lu2022dpmsolver}                                    &  6.61  & 0.65 \\
CT (2-step)~\cite{song2023consistency}                               & 11.10  & 0.56 \\
CT (1-step)~\cite{song2023consistency}                               & 13.00  & 0.47 \\
iCT (1-step)~\cite{song2024ict}                                      &  4.02  & 0.63 \\
StyleGAN2~\cite{karras2020stylegan2}                                 & 21.32  & 0.36 \\
StyleGAN2+GANdance~\cite{kim2024gandance}                            & 17.80  & 0.54 \\
LDHF-DDPM~\cite{rare}                                                &  5.6960 & 0.6811 \\
Proposed $(a=0.5,b=0.5)$                               & \textbf{4.2337} & \textbf{0.6703} \\
Proposed $(a=0.5,b=1.0)$                               & \textbf{3.0429} & \textbf{ 0.6515} \\

\hline
\end{tabular}
\vspace{1.0em}
\caption{Comparison of baseline and proposed methods on the ImageNet dataset (64$\times$64). 
Values for PD, CD, DDIM, DPM-Solver, CT, iCT, StyleGAN2, and StyleGAN2+GANdance are taken from the GANdance ImageNet 64$\times$64 results~\cite{kim2024gandance}; other rows are computed from our implementation using the OpenAI guided-diffusion evaluation pipeline~\cite{ddpm}.}
\label{tab:15}
\end{table}

\subsection{Evaluation on ImageNet 256$\times$256}
\begin{figure}[H]
    \centering
    \includegraphics[width=\textwidth]{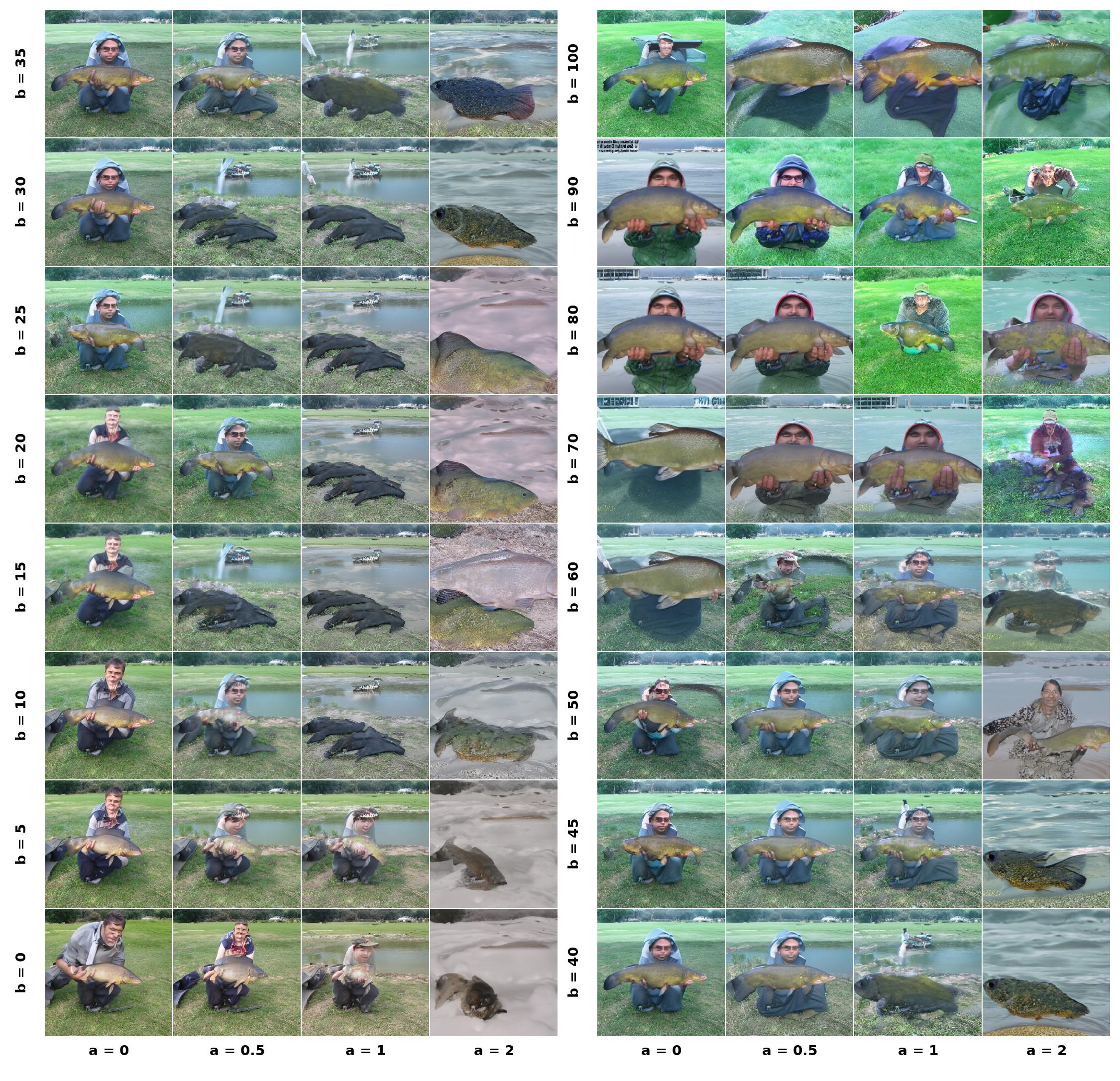}
    \caption{Class "Tench" image. With guidance to generate high-fidelity, diverse images}
    \label{fig:01}
\end{figure}

\begin{figure}[H]
    \centering
    \includegraphics[width=\textwidth]{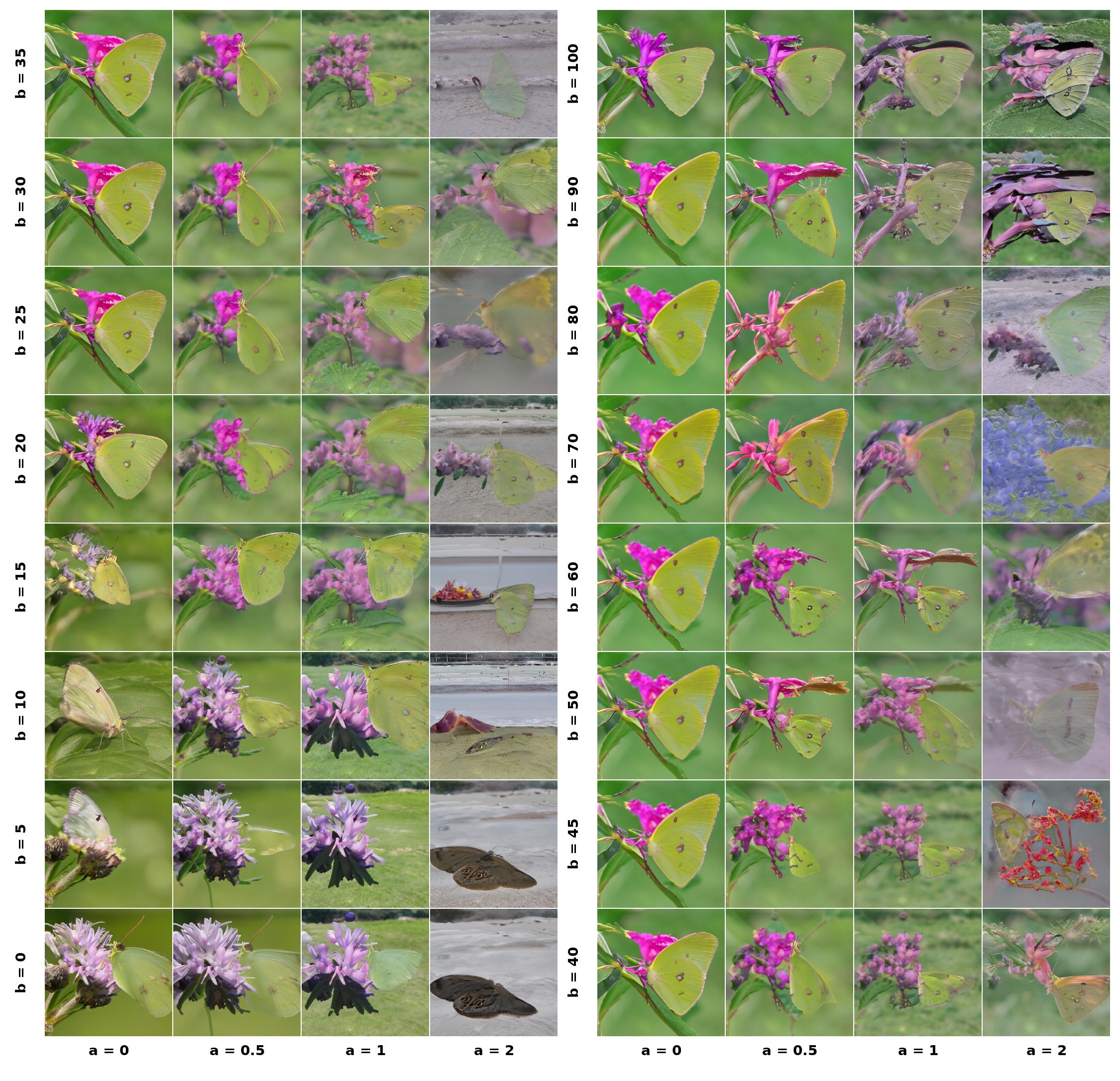}
    \caption{Class "Sulphur Butterfly" image.  With guidance to generate high-fidelity, diverse images}
    \label{fig:13}
\end{figure}

\begin{figure}[H]
    \centering
    \includegraphics[width=\textwidth]{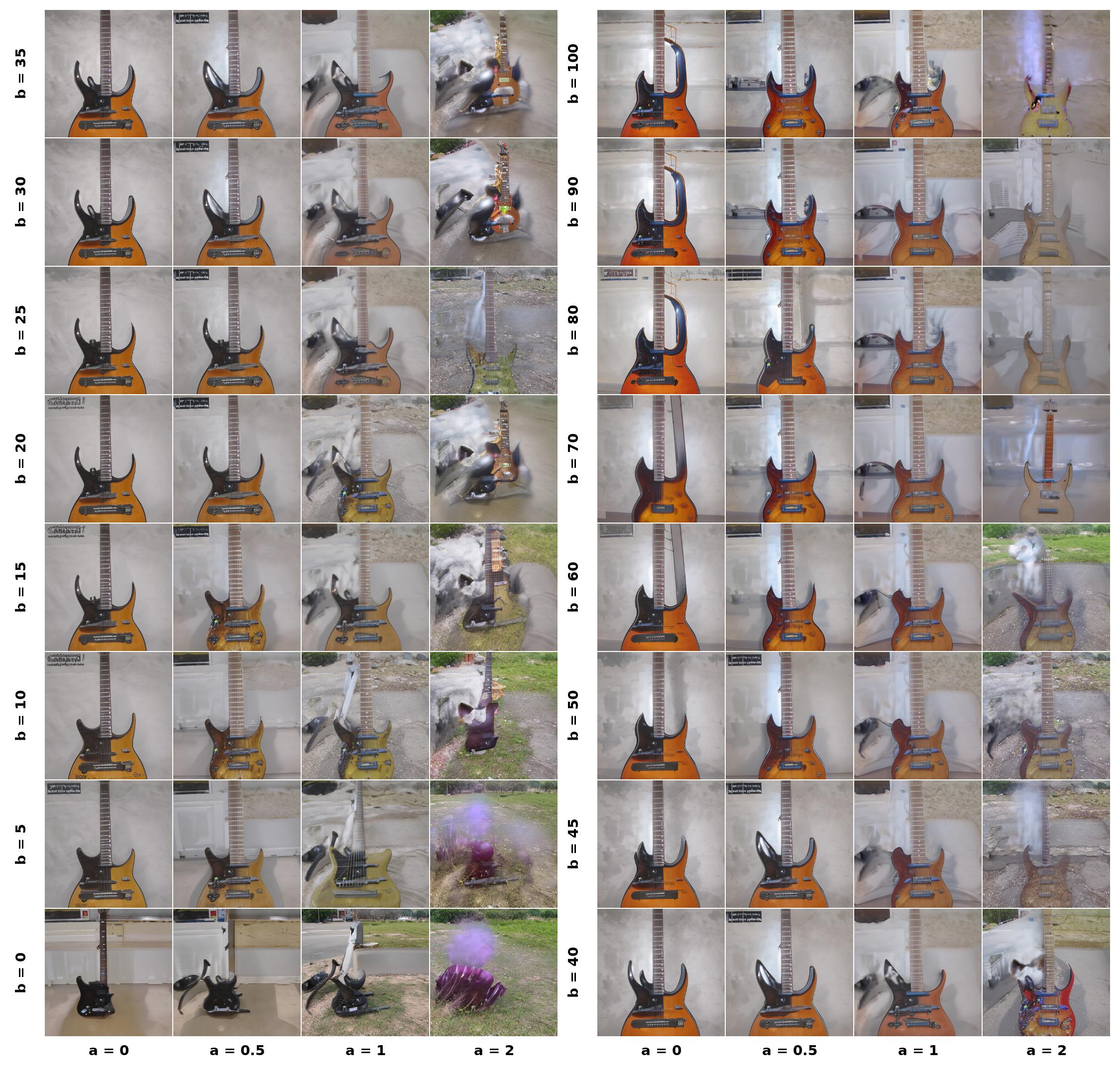}
    \caption{Class "Electric Guitar" image.  With guidance to generate high-fidelity, diverse images}
    \label{fig:21}
\end{figure}

To visualize the effect of both guidance $\nabla_{x_t} \log p_\phi(y_{base}\mid x_t)  $ and $\nabla_{\hat{x}_0}\log p_\phi(y_{base} \mid \hat{x}_0)$ on higher resolution images, we have chosen 3 random classes (Tench, Sulphur butterfly, Electric guitar) of ImageNet and generated 2 images of each class with a fixed seed 42. Then apply our proposed method mentioned in algorithm~\ref{alg:diffusion_sampling} with $a \in {1,2,3,4,5}$ and $b \in {1,2,....,100}$. Generating 50000 samples for each configuration on ImageNet 256x256 for FID and Recall calculation is computationally expensive. That's why we visually show the effect of both guidance on limited images. Moving in the tail direction of the classifier scaled with $a$ leads to generating low possible samples with low fidelity, but by adding additional guidance with the help of $\hat{x}_0$, we move the generation trajectory towards the real data distribution, which is also scaled by $\beta$. For high values of $b$, generated images have high perceptual quality. We have shown the results in figure ~\ref{fig:01},~\ref{fig:13} and in ~\ref{fig:21}. In the given figure, the horizontal line shows the value of $a$ and the vertical line shows the value of $b$. $b$ has only positive values as we move only in the mean direction of the classifier with the help of $\nabla_{\hat{x}_0}\log p_\phi(y_{base} \mid \hat{x}_0)$ guidance. We have applied classifier-based guidance with non-normalized scores. Through experiments, we found that when we scale guidance towards the classifier mean, generated images were more focused on the classified object, and that object size becomes larger. 

\begin{figure}[H]
    \centering
    \includegraphics[width=\textwidth]{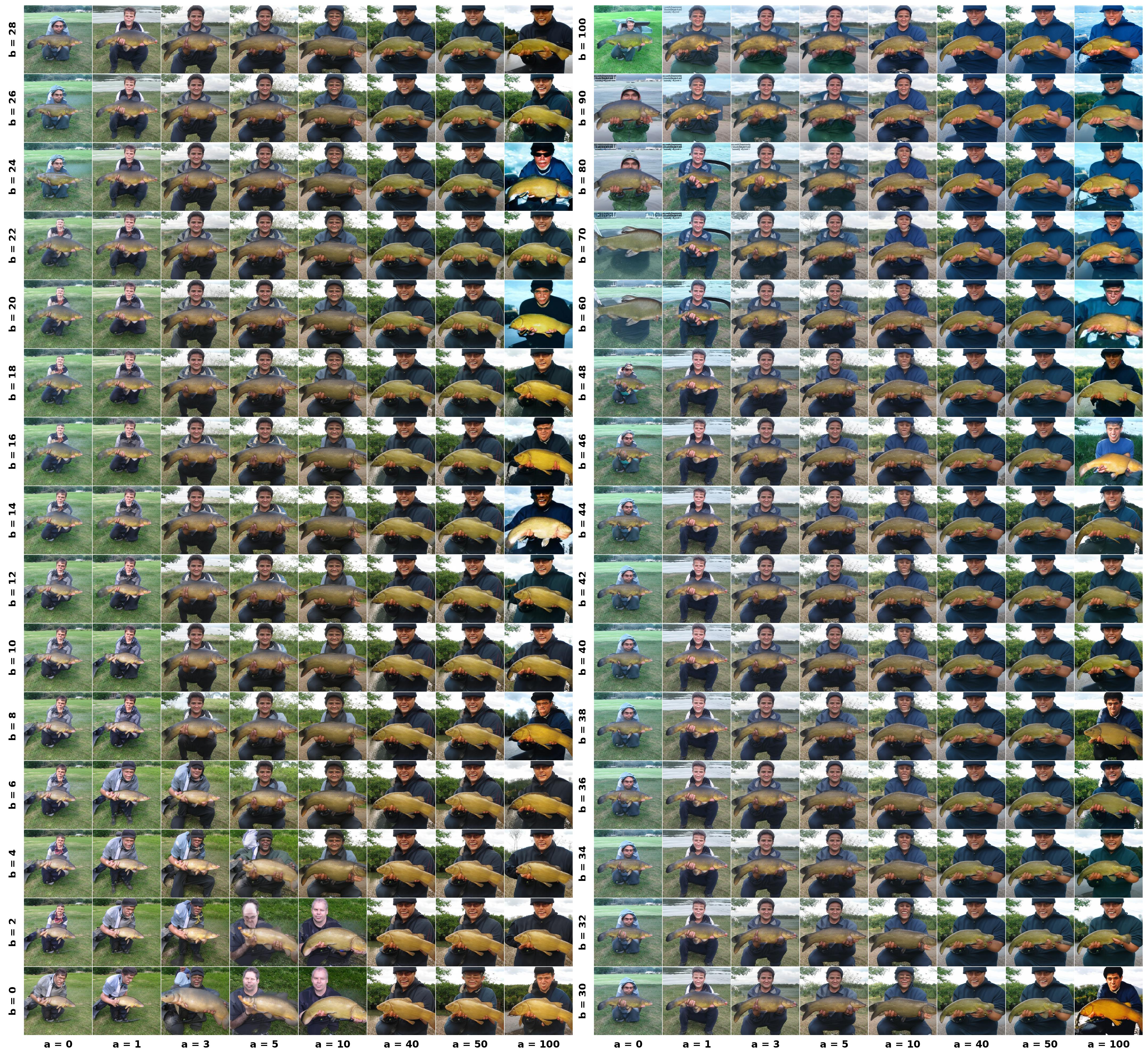}
    \caption{class "tench" image. With both guidance towards the classifier mean to generate high perceptual quality images}
    \label{fig:01c}
\end{figure}

\begin{figure}[H]
    \centering
    \includegraphics[width=\textwidth]{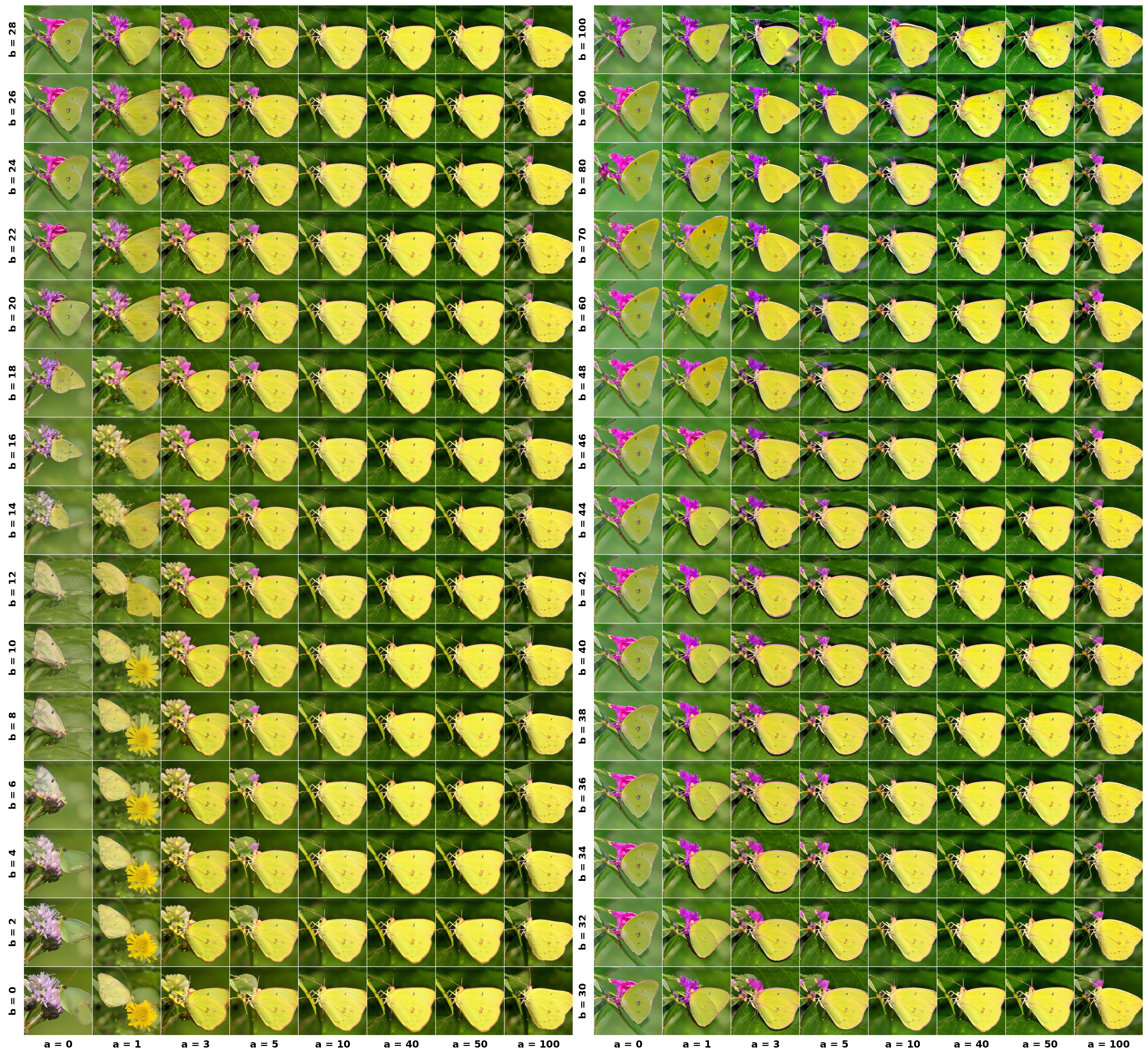}
    \caption{class "Sulphur Butterfly" image. With both guidance towards the classifier mean to generate high perceptual quality images}
    \label{fig:13c}
\end{figure}

\begin{figure}[H]
    \centering
    \includegraphics[width=\textwidth]{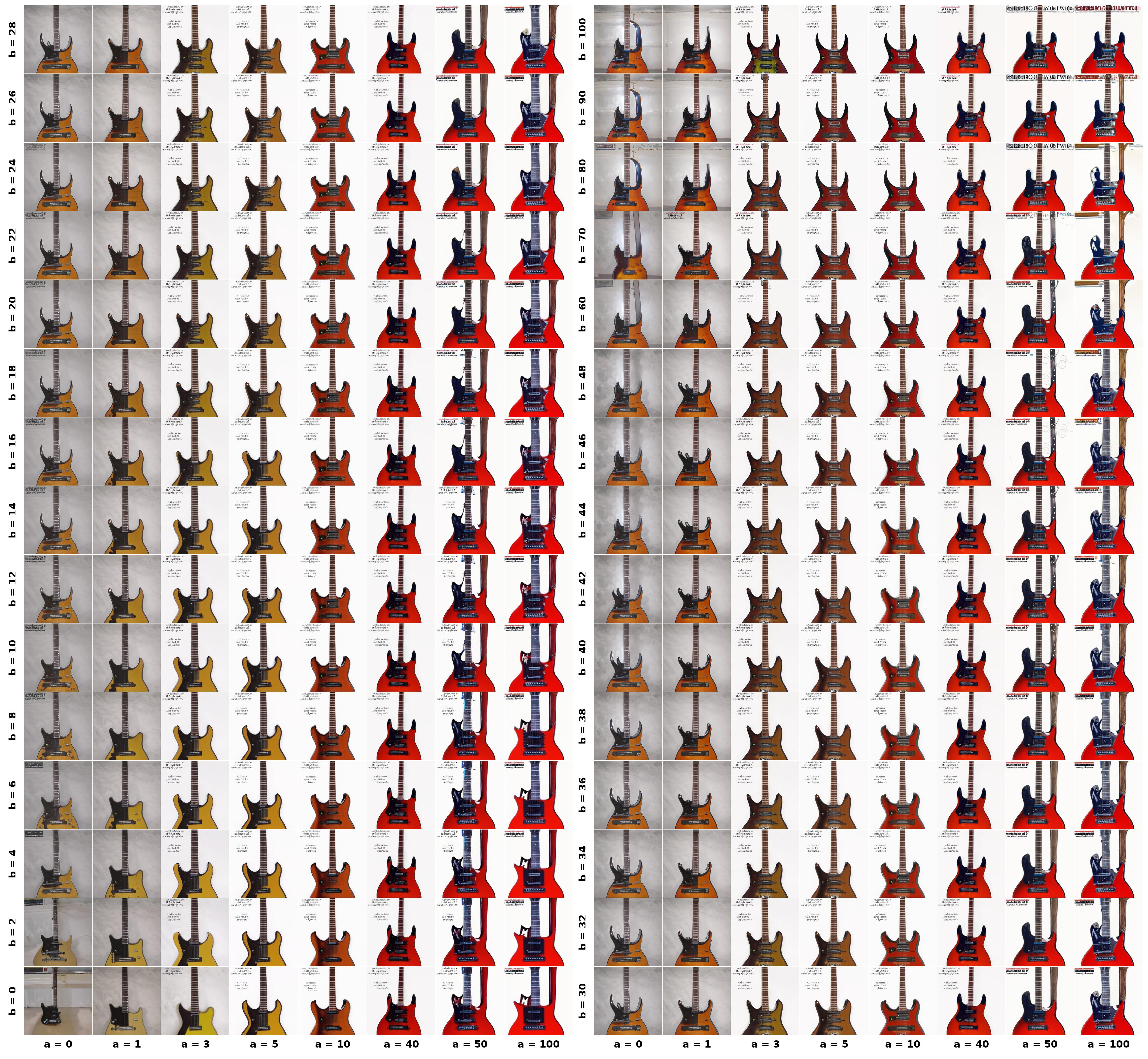}
    \caption{class "Electric Guitar" image. With both guidance towards the classifier mean to generate high perceptual quality images}
    \label{fig:21c}
\end{figure}

We also used this new proposed guidance,$\nabla_{\hat{x}_0}\log p_\phi(y_{base} \mid \hat{x}_0) $along with the original ADM-G guidance $\nabla_{x_t} \log p_\phi(y_{base}\mid x_t)$, where both guidance scores are not normalized, to enhance the quality of generated images. When we combined both guidance in the same direction of the classifier mean, and scaling with positive values $a$ and $b$ as mentioned in algorithm~\ref{alg:diffusion_sampling_2}, the generated images contain fine details of the classified object. Results are shown in figure ~\ref{fig:01c},~\ref{fig:13c} and in ~\ref{fig:21c}. In figure ~\ref{fig:01c}, human hand fingers are now more clear compare to ADM.  In figure ~\ref{fig:13c}, Butterfly fine details are now more perceptually visible. In figure ~\ref{fig:21c}, the electric guitar has finer details.


\section{Conclusion}  

Our experiments indicate that classifier guidance is a key mechanism for controlling the trade-off between fidelity and diversity in class-conditional diffusion models. When the classifier gradient is naively redirected toward low-probability regions of the class posterior, the sampler indeed explores rare intra-class configurations, but this comes at the cost of substantial degradation in image quality. By embedding this low density sample generation guidance $ -\nabla_{x_t} \log p_\phi(y_{base}\mid x_t)$ with $\nabla_{\hat{x}_0}\log p_\phi(y_{base} \mid \hat{x}_0)$ guidance into a variance-weighted, normalized/not-normalized update and scaling it with high values, we obtain a density-aware sampler that simultaneously maintains high fidelity while improving recall in low-density regions. Proposed additional guidance $\nabla_{\hat{x}_0}\log p_\phi(y_{base} \mid \hat{x}_0)$ with original ADM-G guidance helps to generate high-perceptual-quality images.

A central advantage of the proposed approach is that it operates entirely at sampling time: the diffusion model and the classifier remain fixed, and no additional synthetic-vs-real discriminator or tail-specific classifier is required to keep generated samples near the data manifold. The resulting sampler approximates the recall of LDHF-DDPM on the ImageNet dataset while achieving better FID. In future work, we plan to investigate adaptive schedules for the mixing coefficients and guidance strength, extend the analysis to higher resolution samplers, and study how density-aware guidance interacts with more advanced diffusion sampling schemes. A deeper theoretical understanding of how classifier-guided dynamics shape coverage of rare modes is another promising direction.


\begin{thebibliography}{00}

\bibitem{ddpm}
P.~Dhariwal and A.~Nichol, ``Diffusion models beat GANs on image synthesis,'' in \textit{Advances in Neural Information Processing Systems (NeurIPS)}, 2021, vol.~34, pp. 8780--8794.

\bibitem{ho2020ddpm}
J.~Ho, A.~Jain, and P.~Abbeel, ``Denoising diffusion probabilistic models,'' in \textit{Advances in Neural Information Processing Systems (NeurIPS)}, 2020, vol.~33, pp. 6840--6851.

\bibitem{imagenet}
J.~Deng, W.~Dong, R.~Socher, L.~Li, K.~Li, and L.~Fei-Fei, 
``ImageNet: A large-scale hierarchical image database,'' in \textit{Proc. IEEE Conf. Computer Vision and Pattern Recognition (CVPR)}, 2009, pp. 248--255.

\bibitem{rare}
V.~Sehwag, C.~Haz{\i}rbas, A.~Gordo, F.~Ozgenel, and C.~Canton Ferrer, ``Generating high fidelity data from low-density regions using diffusion models,'' in \textit{Proc. IEEE/CVF Conf. Computer Vision and Pattern Recognition (CVPR)}, 2022, pp. 11482--11491.

\bibitem{zhang2024longtail}
T.~Zhang, H.~Zheng, J.~Yao, X.~Wang, M.~Zhou, Y.~Zhang, and Y.~Wang, ``Long-tailed diffusion models with oriented calibration,'' in \textit{Proc. Int. Conf. Learning Representations (ICLR)}, 2024.

\bibitem{ltbsolver2025}
S.~Fu, X.~Liu, and Y.~Wang, ``LTB-Solver: Long-tailed bias solver for image synthesis of diffusion models,'' \textit{Neurocomputing}, vol.~599, 2025.

\bibitem{das2025longdiff}
P.~Das, K.~Fu, A.~Hashemi, and V.~Gupta, ``Principled long-tailed generative modeling via diffusion models,'' in \textit{Advances in Neural Information Processing Systems (NeurIPS)}, 2025.

\bibitem{deligan}
S.~Gurumurthy, R.~K.~Sarvadevabhatla, and R.~Venkatesh Babu, ``DeLiGAN: Generative adversarial networks for diverse and limited data,'' in \textit{Proc. IEEE Conf. Computer Vision and Pattern Recognition (CVPR)}, 2017, pp. 4941--4949.

\bibitem{adagan}
I.~Tolstikhin, S.~Gelly, O.~Bousquet, C.-J.~Simon-Gabriel, and B.~Sch{\"o}lkopf, ``AdaGAN: Boosting generative models,'' in \textit{Advances in Neural Information Processing Systems (NeurIPS)}, 2017, vol.~30, pp. 5424--5433.

\bibitem{ilgan}
J.~R.~Baldvinsson, M.~Ganjalizadeh, A.~AlAbbasi, M.~Bj{\"o}rkman, and A.~H.~Payberah, ``IL-GAN: Rare sample generation via incremental learning in GANs,'' in \textit{Proc. IEEE Global Communications Conf. (GLOBECOM)}, 2022, pp. 621--626.

\bibitem{lee2025divrare}
S.~Lee, J.~Han, S.~Kim, and J.~Choi, ``Diverse rare sample generation with pretrained GANs,'' in \textit{Proc. AAAI Conf. Artificial Intelligence (AAAI)}, 2025, vol.~39, no.~5, pp. 4553--4561.

\bibitem{khorram2024utlo}
S.~Khorram, M.~Jiang, M.~Shahbazi, M.~H.~Danesh, and L.~Fuxin, ``Taming the tail in class-conditional GANs: Knowledge sharing via unconditional training at lower resolutions,'' in \textit{Proc. IEEE/CVF Conf. Computer Vision and Pattern Recognition (CVPR)}, 2024, pp. 7580--7590.

\bibitem{rangwani2021cbgan}
H.~Rangwani, K.~R.~Mopuri, and R.~V.~Babu, ``Class balancing GAN with a classifier in the loop,'' in \textit{Proc. Int. Conf. Learning Representations (ICLR)}, 2021.

\bibitem{schreurs2021leverage}
J.~Schreurs, H.~De~Meulemeester, M.~Fanuel, B.~De~Moor, and J.~A.~K.~Suykens, ``Leverage score sampling for complete mode coverage in generative adversarial networks,'' in \textit{Machine Learning and Knowledge Discovery in Databases}, 2021, pp. 593--609.

\bibitem{turner2019mhgan}
R.~Turner, J.~Hung, E.~Frank, Y.~Saatci, and J.~Yosinski, ``Metropolis-Hastings generative adversarial networks,'' in \textit{Proc. Int. Conf. Machine Learning (ICML)}, 2019, pp. 6345--6353.

\bibitem{heusel2017fid}
M.~Heusel, H.~Ramsauer, T.~Unterthiner, B.~Nessler, and S.~Hochreiter, ``GANs trained by a two time-scale update rule converge to a local Nash equilibrium,'' in \textit{Advances in Neural Information Processing Systems (NeurIPS)}, 2017, vol.~30, pp. 6626--6637.

\bibitem{kynkaanniemi2019pr}
T.~Kynk{\"a}{\"a}nniemi, T.~Karras, S.~Laine, J.~Lehtinen, and T.~Aila, ``Improved precision and recall metric for assessing generative models,'' in \textit{Advances in Neural Information Processing Systems (NeurIPS)}, 2019, vol.~32.

\bibitem{naeem2020prdc}
M.~F.~Naeem, S.~J.~Oh, Y.~Uh, Y.~Choi, and J.~Yoo, ``Reliable fidelity and diversity metrics for generative models,'' in \textit{Proc. Int. Conf. Machine Learning (ICML)}, 2020, pp. 7133--7142.

\bibitem{radford2021clip}
A.~Radford \textit{et al.}, ``Learning transferable visual models from natural language supervision,'' in \textit{Proc. Int. Conf. Machine Learning (ICML)}, 2021, pp. 8748--8763.

\bibitem{szegedy2016inception}
C.~Szegedy, V.~Vanhoucke, S.~Ioffe, J.~Shlens, and Z.~Wojna, ``Rethinking the Inception architecture for computer vision,'' in \textit{Proc. IEEE Conf. Computer Vision and Pattern Recognition (CVPR)}, 2016, pp. 2818--2826.

\bibitem{zhang2018lpips}
R.~Zhang, P.~Isola, A.~A.~Efros, E.~Shechtman, and O.~Wang, ``The unreasonable effectiveness of deep features as a perceptual metric,'' in \textit{Proc. IEEE/CVF Conf. Computer Vision and Pattern Recognition (CVPR)}, 2018, pp. 586--595.


\bibitem{song2019score}
Y.~Song and S.~Ermon,
``Generative modeling by estimating gradients of the data distribution,''
in \emph{Advances in Neural Information Processing Systems (NeurIPS)}, 2019, pp. 11895--11907.

\bibitem{song2021sde}
Y.~Song, J.~Sohl-Dickstein, D.~P.~Kingma, A.~Kumar, S.~Ermon, and B.~Poole,
``Score-based generative modeling through stochastic differential equations,''
in \emph{Proc. Int. Conf. Learning Representations (ICLR)}, 2021.

\bibitem{nichol2021improved}
A.~Q.~Nichol and P.~Dhariwal,
``Improved denoising diffusion probabilistic models,''
in \emph{Proc. Int. Conf. Machine Learning (ICML)}, 2021, pp.~8162--8171.

\bibitem{sohl2015}
J.~Sohl-Dickstein, E.~A.~Weiss, N.~Maheswaranathan, and S.~Ganguli,
``Deep unsupervised learning using nonequilibrium thermodynamics,''
in \emph{Proc. Int. Conf. Machine Learning (ICML)}, 2015, pp.~2256--2265.

\bibitem{mcinnes2018umap}
L.~McInnes, J.~Healy, and J.~Melville,
``UMAP: Uniform Manifold Approximation and Projection for dimension reduction,''
\emph{arXiv:1802.03426}, 2018.

\bibitem{ledoit2004well}
O.~Ledoit and M.~Wolf,
``A well-conditioned estimator for large-dimensional covariance matrices,''
\emph{Journal of Multivariate Analysis}, vol.~88, no.~2, pp.~365--411, 2004.

\bibitem{radford2021clip}
A.~Radford, J.~W. Kim, C.~Hallacy, A.~Ramesh, G.~Goh, S.~Agarwal, G.~Sastry,
  A.~Askell, P.~Mishkin, J.~Clark, G.~Krueger, and I.~Sutskever,
``Learning transferable visual models from natural language supervision,''
in \emph{Proc. Int. Conf. Machine Learning (ICML)}, 2021, pp.~8748--8763.

\bibitem{zhang2018lpips}
R.~Zhang, P.~Isola, A.~A. Efros, E.~Shechtman, and O.~Wang,
``The unreasonable effectiveness of deep features as a perceptual metric,''
in \emph{Proc. IEEE/CVF Conf. Computer Vision and Pattern Recognition (CVPR)}, 2018, pp.~586--595.

\bibitem{szegedy2016inception}
C.~Szegedy, V.~Vanhoucke, S.~Ioffe, J.~Shlens, and Z.~Wojna,
``Rethinking the Inception architecture for computer vision,''
in \emph{Proc. IEEE Conf. Computer Vision and Pattern Recognition (CVPR)}, 2016, pp.~2818--2826.

\bibitem{salimans2022progressivedistillation}
T.~Salimans and J.~Ho,
``Progressive distillation for fast sampling of diffusion models,''
in \emph{Proc. Int. Conf. Learning Representations (ICLR)}, 2022.

\bibitem{song2023consistency}
Y.~Song, P.~Dhariwal, M.~Chen, and I.~Sutskever,
``Consistency models,''
in \emph{Proc. Int. Conf. Machine Learning (ICML)}, 2023.

\bibitem{karras2022edm}
T.~Karras, M.~Aittala, S.~Laine, and T.~Aila,
``Elucidating the design space of diffusion-based generative models,''
in \emph{Advances in Neural Information Processing Systems (NeurIPS)}, 2022.

\bibitem{song2021ddim}
J.~Song, C.~Meng, and S.~Ermon,
``Denoising diffusion implicit models,''
in \emph{Proc. Int. Conf. Learning Representations (ICLR)}, 2021.

\bibitem{lu2022dpmsolver}
C.~Lu, Y.~Zhou, F.~Bao, J.~Chen, C.~Li, and J.~Zhu,
``DPM-Solver: A fast ODE solver for diffusion probabilistic model sampling in around 10 steps,''
in \emph{Advances in Neural Information Processing Systems (NeurIPS)}, 2022.

\bibitem{song2024ict}
Y.~Song and P.~Dhariwal,
``Improved techniques for training consistency models,''
in \emph{Proc. Int. Conf. Learning Representations (ICLR)}, 2024.

\bibitem{karras2020stylegan2}
T.~Karras, S.~Laine, M.~Aittala, J.~Hellsten, J.~Lehtinen, and T.~Aila,
``Analyzing and improving the image quality of StyleGAN,''
in \emph{Proc. IEEE/CVF Conf. Computer Vision and Pattern Recognition (CVPR)}, 2020, pp.~8107--8116.

\bibitem{kim2024gandance}
J.~Kim \emph{et al.},
``GANdance: Guided sampling for conditional generative adversarial networks,''
in \emph{Proc. Eur. Conf. Computer Vision (ECCV)}, 2024.


\end{thebibliography}
\end{document}